%% file: acl_latex.tex
\pdfoutput=1

\documentclass[11pt]{article}

\usepackage{acl}

\usepackage{times}
\usepackage{latexsym}
\usepackage{graphicx}
\usepackage{enumitem}

\usepackage[T1]{fontenc}
\usepackage{tipa}

\usepackage[utf8]{inputenc}

\usepackage{microtype}

\usepackage{booktabs}
\usepackage{xcolor}
\definecolor{boxgray}{HTML}{FAFAFA}
\definecolor{boxgray2}{HTML}{C0C0C0}
\newcommand{\ba}{`}

\title{User-Centric Evaluation of OCR Systems for Kwak'wala}

\author{
    Shruti Rijhwani,\textsuperscript{$1$} Daisy Rosenblum,\textsuperscript{$2$} Michayla King,\textsuperscript{$3$} \\ \textbf{Antonios Anastasopoulos,\textsuperscript{$4$} Graham Neubig\textsuperscript{$1$} } \\
    \textsuperscript{$1$}Language Technologies Institute, Carnegie Mellon University\\
    \textsuperscript{$2$}University of British Columbia\\
    \textsuperscript{$3$}\b K’w\b ala Language Program\\
    \textsuperscript{$4$}Department of Computer Science, George Mason University\\
  \texttt{srijhwan@cs.cmu.edu, daisy.rosenblum@ubc.ca, michayla.g3@gmail.com}\\ \texttt{antonis@gmu.edu, gneubig@cs.cmu.edu}
}

\begin{document}
\maketitle
\begin{abstract}
 There has been recent interest in improving optical character recognition (OCR) for endangered languages, particularly because a large number of documents and books in these languages are not in machine-readable formats. The performance of OCR systems is typically evaluated using automatic metrics such as character and word error rates. While error rates are useful for the comparison of different models and systems, they do not measure whether and how the transcriptions produced from OCR tools are useful to downstream users. In this paper, we present a human-centric evaluation of OCR systems, focusing on the Kwak'wala language as a case study. With a user study, we show that utilizing OCR reduces the time spent in the manual transcription of culturally valuable documents -- a task that is often undertaken by endangered language community members and researchers -- by over 50\%. Our results demonstrate the potential benefits that OCR tools can have on downstream language documentation and revitalization efforts.\footnote{Code, models, and datasets are available at \url{https://shrutirij.github.io/ocr-el/}.}
\end{abstract}

\section{Introduction}

Documentation and revitalization efforts for endangered languages frequently lead to the creation of textual documents in these languages. These include cultural materials such as folk tales and poetry; linguistic documentation like speech transcriptions and vocabulary lists; and other archival material~\citep{himmelmann1998documentary,grenoble2005saving}. However, even though a substantial number of such documents have been created for endangered languages around the globe, the vast majority are not widely accessible because they exist only as printed books and handwritten notes.

Although some of these documents are digitally available as scanned images, the text contained in the images is not machine-readable, inhibiting several use cases that are important to communities that speak endangered languages. For example, (1) the text is not searchable for speakers and researchers of these languages; (2) it cannot be reformatted, indexed, or adapted to various needs; and (3) it cannot be used to build datasets for training NLP models. Machine-readable transcriptions of documents are typically produced by a human transcriber, who looks at the document and retypes the text present in it. Like other manual transcription tasks (e.g., speech transcription), this process is time-consuming and requires significant effort. 

That said, there are computational approaches to producing machine-readable text from scanned documents, specifically through optical character recognition (OCR).  Training a high-performance OCR system is challenging given the small amount of data that is typically available in endangered languages. However, there has been recent interest~\cite{rijhwani-etal-2020-ocr, rijhwani-etal-2021-lexically, tjuatjaexplorations, pintupi} in improving OCR even in very low-resourced settings using the technique of \textit{automatic post-correction}. Post-correction models correct errors in existing OCR transcriptions~\cite{kolak-resnik-2005-ocr,dong-smith-2018-multi, krishna-etal-2018-upcycle}. The post-correction methods presented in \citet{rijhwani-etal-2021-lexically} demonstrated substantial performance gains for multiple low-resourced endangered languages -- reducing character error rates (CER) by 32--58\% and word error rates (WER) by 29--59\% relative to off-the-shelf OCR systems.\footnote{Character error rate (CER) and word error rate (WER) are based on edit distance and are standard metrics for evaluating OCR systems~\cite{berg-kirkpatrick-etal-2013-unsupervised,schulz-kuhn-2017-multi}. CER is the edit distance between the predicted and the gold transcriptions of the document, divided by the total number of characters in the gold transcription. WER is similar but is calculated at the word level.}

While error rates are useful to quantify the performance of various OCR technologies, they do not measure whether the produced transcriptions are useful to the primary audience for these transcriptions: community language learners, teachers, and researchers. In this paper, we look beyond error rates and take a human-centered approach to evaluating OCR and understanding whether the automatically produced transcriptions are beneficial to downstream users. More specifically, we analyze whether OCR is effective in lowering the time and effort spent in manually creating accurate transcriptions of scanned documents which, as discussed above, is a task that is frequently undertaken in language documentation and preservation programs.

As a case study, we focus on Kwak'wala, an endangered language spoken in North America, because of its long tradition of written documentation and active community engagement in accessing the knowledge contained in these texts (detailed in \autoref{sec:docs}). We conduct a user study where we compare the time spent by human transcribers on producing an accurate transcription of typewritten Kwak'wala documents with and without the use of an OCR system.\footnote{Similar user studies have been carried out to determine the effectiveness of machine translation in reducing human post-editing effort~\cite{specia2010estimating,gaspari-etal-2014-perception,koponen2016machine}, but none for OCR or endangered languages, to the best of our knowledge. \citet{kettunen2022ocr} measure user-perceived (qualitative) utility of OCR transcripts based on information gain, as opposed to our quantitative study on reducing transcription time.} 
We demonstrate that there is a statistically significant reduction in the time needed for manual transcription when an OCR system is used beforehand. Our results indicate that further research and development of improved OCR tools for endangered languages can add valuable efficiency to language preservation and revitalization efforts.

\section{Documents in the Kwak'wala Language}
\label{sec:docs}

To conduct our proposed human-centric evaluation of OCR, we focus on documents in the Kwak'wala language, while noting that our user study does not involve any language-specific components and can be extended to other languages.

Kwak'wala is a member of the Wakashan language family spoken on the Northwest North American Coast. Heritage learners and teachers are actively engaged in the revitalization of Kwak’wala.
Written documentation of the language extends back over 120 years, including a collection of documents produced by anthropologist Franz Boas in collaboration with George Hunt, a native speaker of Kwak'wala~\cite[\textit{inter alia}]{boas1897social, boas1902, boas1911, boas1921, boas1934}. The Hunt-Boas documents include 14 published volumes and several more unpublished manuscripts. The documents encompass a grammar of the language; word lists; stories; recipes; procedural texts; descriptions of practices, beliefs, and customs; descriptions of dialectal differences; maps and lists of placenames; and more. For Kwak'wala communities and language researchers today, these texts are rich troves containing knowledge that has special value to community-led projects focused on teaching, learning, strengthening, and reclaiming their language, cultural practices, and territorial sovereignty~\citep{Lawson_2004}.

\begin{figure}[t]
    \centering
    \includegraphics[width=0.48\textwidth]{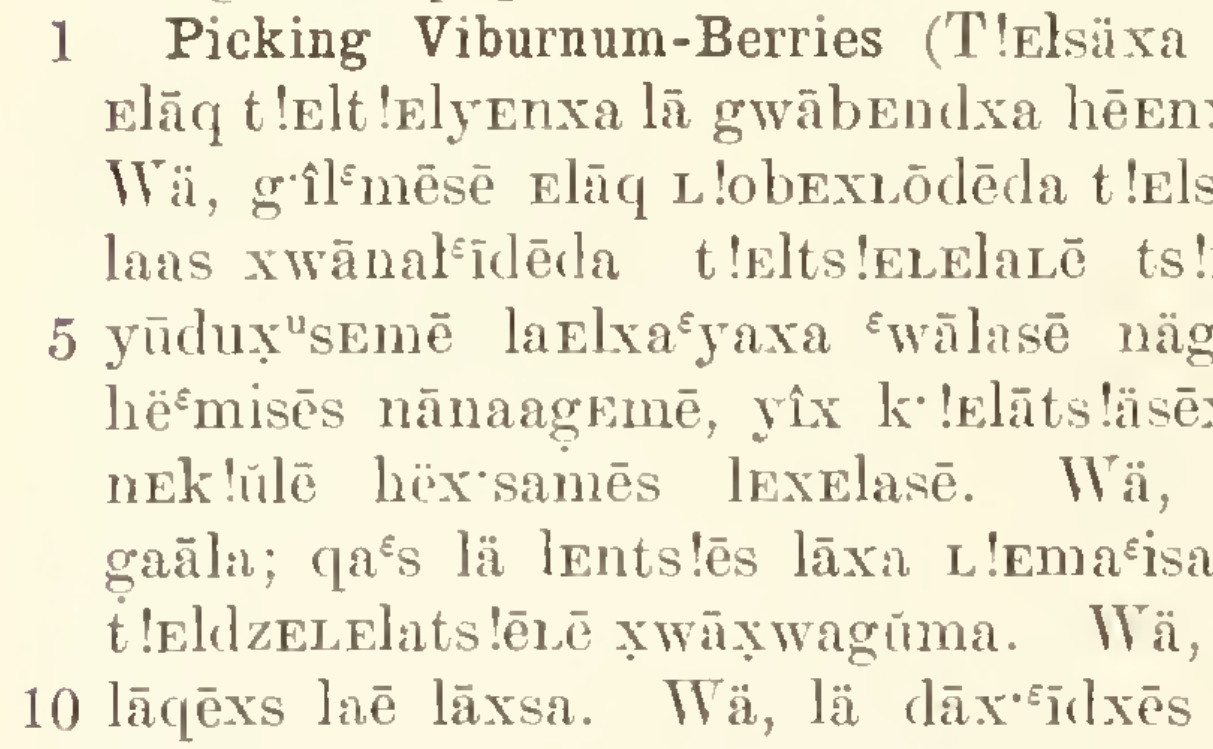}
    \caption{An excerpt from the Hunt-Boas publications documenting the community's method for picking viburnum berries. As seen, the Hunt-Boas orthography is complex -- it uses several digraphs and diacritics that are challenging for an OCR system to recognize.}
    \label{fig:boas_example}
\end{figure}
However, to the extent the Hunt-Boas documents have been digitized, they are still \ba trapped' in scanned images. The texts are not searchable and researchers potentially need to look at tens or hundreds of images to locate relevant information. Moreover, the Hunt-Boas orthography is technical and somewhat idiosyncratic and is primarily used in archival research contexts -- because the texts are not machine-readable, they cannot be automatically transliterated to modern, community-preferred orthographies. Researchers who draw on these materials often resort to retyping excerpts (sometimes into a different writing system), a time-consuming process that introduces a tight bottleneck to sharing and accessing this knowledge. 

Therefore, extracting the Hunt-Boas texts into a machine-readable format can serve the community in many ways. Our user study, thus, focuses on evaluating the utility of existing OCR techniques as applied to these culturally important documents. We select OCR systems based on the experiments in \citet{rijhwani-etal-2021-lexically} which describe two models that worked particularly well on the challenging Hunt-Boas orthography (\autoref{fig:boas_example} has an example):

\begin{itemize}
\item \textbf{Ocular} is an unsupervised OCR system that uses a generative model to transcribe scanned documents~\cite{berg-kirkpatrick-etal-2013-unsupervised,garrette-etal-2015-unsupervised}. Ocular’s transcription model relies on a character n-gram language model trained on the target language. \citet{rijhwani-etal-2021-lexically} use a small amount of Kwak'wala text data to train the language model and show that Ocular's OCR system resulted in a CER of 7.90\% and a WER of 38.22\% on the Hunt-Boas texts.
\item \textbf{Post-correction} involves correcting the errors made by an existing OCR system to improve overall accuracy. \citet{rijhwani-etal-2021-lexically} present a neural encoder-decoder model~\cite{bahdanau2015neural} trained with semi-supervised learning to improve post-correction performance in low-resource scenarios. Relative to Ocular, the post-correction method reduces the CER by 52\% and the WER by 41\% on the Kwak'wala data.
\end{itemize}

In the following sections, we describe a user study focused on evaluating the two OCR pipelines (Ocular and post-correction) to understand whether the automatically produced transcriptions are beneficial to downstream users that access the information in the Hunt-Boas publications.

\section{Evaluation with a User Study}
\label{sec:userstudy}
Traditionally, accurate transcriptions of the Hunt-Boas documents are produced by a human transcriber (often a Kwak'wala community member, linguistic researcher, or archivist). The transcriber looks at the scanned image of each document and types out the text present in it -- a time-consuming process. To evaluate the utility of the outputs from OCR models, we conduct a user study where we compare the time spent by transcribers on producing an accurate transcription in various settings with and without the use of an OCR system. We attempt to answer two primary questions: 

\begin{enumerate}
    \item Is it faster for a human transcriber to correct the errors in an OCR output as compared to typing out the text from scratch?
    \item Does adding a post-correction model affect transcription speed beyond existing off-the-shelf OCR tools such as Ocular?
\end{enumerate}

We design controlled experiments to measure human transcription speed on a subset of images from the Hunt-Boas texts and evaluate how the speed is affected in various settings to understand whether there is utility in introducing OCR into the process. Additionally, we obtain subjective feedback on how having OCR outputs affected the transcription task through a survey sent to participating transcribers after tasks were completed. 

\subsection{Participants}
We employed nine participants for the user study, all of whom had some transcription experience. Of the nine, two participants had familiarity with the Kwak’wala language as well as the Hunt-Boas texts and the orthography -- one is a heritage Kwak’wala language learner and the other is an academic linguist working with Kwak’wala language materials.

We also employed seven participants that had no experience or familiarity with Kwak’wala. Three of these participants are computer science graduate students at a university and four participants were employed through Upwork,\footnote{\url{https://www.upwork.com/}} a marketplace for freelance professionals. We selected them based on prior transcription experience, knowledge about data annotation for machine learning, and linguistic training as well as a high job success rate on the Upwork platform.\footnote{Full IRB approval was obtained for the user study; all participants signed a consent form before working on the transcription tasks; and all data collected was anonymized.}
Including participants with varying degrees of prior knowledge of the Kwak’wala language also allowed us to evaluate whether this is a factor that affects transcription speed and the overall experience with the user study tasks.

\subsection{Transcription Interface and Keyboard}

\begin{figure*}[t]
    \centering
    \fcolorbox{boxgray2}{boxgray}{\includegraphics[width=0.9\textwidth]{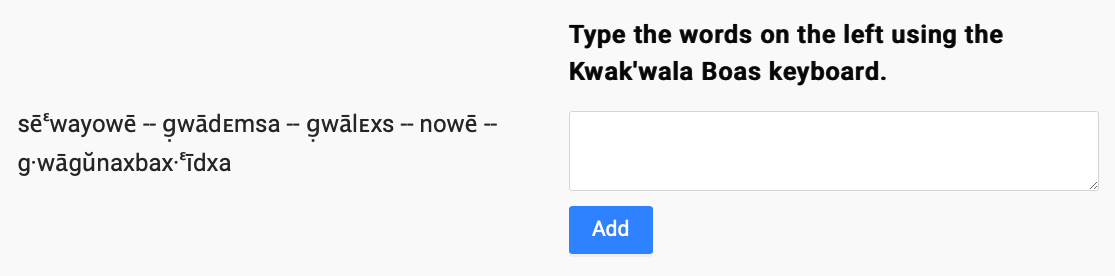}}
    \caption{Practice task for transcribers to become familiar with the Boas keyboard. We included eight practice tasks in the Label Studio interface to cover all special character combinations in the Boas orthography multiple times. Users could repeat tasks as many times as they wanted to before moving on to the main transcription task.}
    \label{fig:keyboard_task}
\end{figure*}

\begin{figure*}[t]
    \centering
    \fcolorbox{boxgray2}{boxgray}{\includegraphics[width=0.98\textwidth]{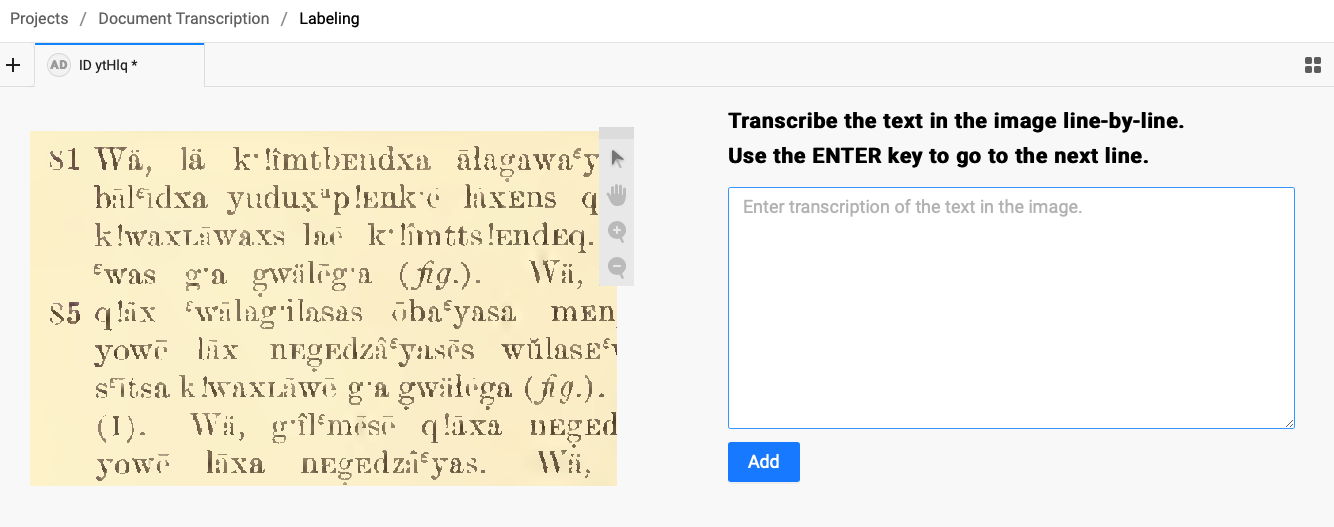}}
    \caption{Transcription task interface, designed in Label Studio. The interface displays the image of a page and a text box to enter the transcription. It also has zoom and pan tools for the image, allowing users to zoom in on characters that might be hard to identify. The figure depicts a cropped image for clarity. When an OCR system is used before the manual transcription task, the text box on the right is pre-filled with the output transcription from the model and the user's task is to correct any remaining errors.}
    \label{fig:transcription_task}
\end{figure*}

We use Label Studio,\footnote{\url{https://labelstud.io}} an open-source data annotation interface for setting up transcription tasks for the user study. We customized the interface for the transcription task and additionally modified it to record information necessary for our analysis of transcription speed, including timestamps for when transcribers operate on each task.

Many characters and diacritics in the Hunt-Boas orthography are not present on a standard computer keyboard. To increase transcription efficiency, we used Keyman Developer\footnote{\url{https://keyman.com/developer/}} (an open-source toolkit) to create a keyboard for representing the characters in the orthography. The keyboard maps standard US English keyboard keystrokes to characters in the Hunt-Boas orthography. A detailed description of the keyboard layout and usage is in \autoref{sec:keyboard}. All participants were required to use this virtual keyboard to ensure consistency in terms of typing efficiency across all transcribers. 

To train participants before the user study experiments, we designed a keyboard practice task, which presents a few sentences of text in the Hunt-Boas orthography that the transcriber has to type using the keyboard. The practice texts were selected such that all the different diacritic and digraph keystroke combinations were covered multiple times. The practice tasks were also added to the Label Studio web interface -- a screenshot of the interface for the practice task is shown in \autoref{fig:keyboard_task}. Participants were able to repeat the practice tasks as many times as needed to gain familiarity with the keyboard. Additionally, we added keystroke mapping information to the interface for all tasks (transcription and practice tasks) for users to quickly reference.

\subsection{Transcription Task Settings}

The primary objective for the participants was to produce an accurate transcription of the image presented to them in each task. In the Label Studio interface, as seen in \autoref{fig:transcription_task}, the image is displayed alongside a text box for the user to enter the transcription. To evaluate whether using OCR is useful in reducing transcription speed, we have three different setups for the tasks:

\begin{itemize}
    \item \textbf{Baseline}: This setup does not include the use of any OCR system. The transcriber must type out the text seen in the image from scratch -- they are presented with the image and an empty text box in the interface (see \autoref{fig:transcription_task}). This setup represents our baseline for measuring transcription speed, as this is the method currently used by Kwak'wala researchers and community members.
    
    \item \textbf{Ocular}: In this setup, we use the off-the-shelf OCR tool Ocular on the image for each task before manual annotation. The transcriber is presented with the image and a text box containing the OCR output -- that is, the text box on the right in \autoref{fig:transcription_task} will be pre-filled with the OCR output. The task here involves looking at the text present in the image (which is the target text) and editing the OCR output in the text box to correct all the errors and produce an accurate transcription.
    
    \item \textbf{Post-correction}: This is similar to the previous setup, but we use a pipeline that includes applying the OCR post-correction method from \citet{rijhwani-etal-2021-lexically}, and as described in \autoref{sec:docs}), it improves OCR performance (CER and WER) on Kwak'wala text as compared to Ocular. The transcriber is presented with the image alongside a text box containing the post-corrected transcription. The task is to correct any remaining errors.
\end{itemize}

\subsection{Experiment Design}
While measuring transcription speed for a single page is relatively straightforward, determining whether there is a statistically significant difference in speed between the three different setups described above requires consideration of several factors. For example, a single transcriber cannot be assigned the same page multiple times with different setups as they would become familiar with the page’s content, potentially leading to incorrect estimation of speed differences. Additionally, some participants may be faster at transcription in general and some pages in the document may be more challenging than others -- these factors need to be accounted for when measuring transcription time across the task setups.

\begin{figure}
    \centering
    \begin{tabular}{|c|c|c|c|}
        \hline
         A & B & C & D \\
         \hline
         B & C & D & A \\
         \hline
         C & D & A & B \\
         \hline
         D & A & B & C \\
         \hline
    \end{tabular}
    \quad
    \begin{tabular}{|c|c|c|c|}
        \hline
         A & B & C & D \\
         \hline
         B & A & D & C \\
         \hline
         C & D & B & A \\
         \hline
         D & C & A & B \\
         \hline
    \end{tabular}
    \caption{Two 4x4 Latin Squares. Each symbol appears only once in each row and each column. The number of symbols is the same as the number of rows and columns. Figure adapted from \citet{dean1999design}.}
    \label{fig:latin_square}
\end{figure}

In statistics, such factors are known as sources of variability (or nuisance factors). We design the transcription tasks to control the variability introduced by these factors using the Latin Square Design~\citep{dean1999design} to assign tasks to each transcriber. The Latin Square has the same number of rows and columns (square-shaped), with a specific symbol appearing exactly once in each row and exactly once in each column; \autoref{fig:latin_square} shows two examples of a Latin Square design that has 4 rows and 4 columns. This design allows control of two sources of variability -- one along the rows and one along the columns.

\begin{figure*}
    \centering
    \begin{tabular}{c|c|c|c|}
         \multicolumn{1}{c}{} & \multicolumn{1}{c}{page1} & \multicolumn{1}{c}{page2} & \multicolumn{1}{c}{page3} \\
         \cline{2-4}
         user1 & ocu & base & post \\
         \cline{2-4}
         user2 & post & ocu & base \\
         \cline{2-4}
         user3 & base & post & ocu \\
         \cline{2-4}
    \end{tabular}
    \quad
    \begin{tabular}{c|c|c|c|}
         \multicolumn{1}{c}{} & \multicolumn{1}{c}{page4} & \multicolumn{1}{c}{page5} & \multicolumn{1}{c}{page6} \\
         \cline{2-4}
          & ocu & post & base \\
         \cline{2-4}
          & base & ocu & post \\
         \cline{2-4}
          & post & base & ocu \\
         \cline{2-4}
    \end{tabular}
    \quad
    \begin{tabular}{c|c|c|c|}
        \multicolumn{1}{c}{} &  \multicolumn{1}{c}{page7} & \multicolumn{1}{c}{page8} & \multicolumn{1}{c}{page9} \\
         \cline{2-4}
          & post & ocu & base \\
         \cline{2-4}
          & base & post & ocu \\
         \cline{2-4}
          & ocu & base & post \\
         \cline{2-4}
    \end{tabular}
    \caption{Task setup assignments for a group of three users using the Latin Square design. We use 3x3 Latin Squares because we have three task setups: Baseline (\textit{base}), Ocular (\textit{ocu}), and post-correction (\textit{post}). We need three squares for each group of users because we have nine pages for transcription. All users transcribe the same set of pages, but with the Latin Square framework, they have different task setups for each page which helps control sources of variability. All user identifiers and page identifiers are randomized before applying the Latin Square design.}
    \label{fig:task_squares}
\end{figure*}

Since we have three task setups, we choose a 3x3 Latin Square -- each setup appears only once in each row and column. The two sources of variability we control are (1) the user doing the transcription and (2) the page being transcribed. We randomly divide the nine participants into three groups of three users each (to fit the 3x3 square) and choose a fixed set of nine pages from the documents that all participants will transcribe in their tasks. For each group of three users, we form three squares (since we have nine pages). The task setups -- i.e., baseline, Ocular, post-correction -- are randomly assigned within the Latin Square constraints. Adding randomization for all factors (user, page, task setup) is aimed at spreading out the effect of undetectable or unsuspected characteristics. An example of task setup assignments for one group of three users for the nine pages is in \autoref{fig:task_squares}.\footnote{We follow \url{https://online.stat.psu.edu/stat503/lesson/4/4.4} and randomize Latin Squares separately for each group of users and each set of pages, so task setup assignments may not look identical across groups.}

Therefore, each user has nine transcription tasks with the task setups evenly distributed so all users are sufficiently timed on each setup. The user does not transcribe the same page more than once, but all users transcribe the same set of nine pages (with varied task setups). The Latin Square Design, thus, introduces randomness across the factors to reduce variance and improve the generalization of the statistical analysis.

\paragraph{Dataset selection} We selected nine pages from the Hunt-Boas volumes for the user study experiments, which were randomly chosen from a larger subset of 50 pages that community-based researchers deemed representative of the volumes and important to transcribe.

\subsection{Evaluation Procedure}
The nine transcription tasks were designed to take approximately 7 hours to complete. The participants accessed the Label Studio interface remotely through any web browser and first completed the keyboard practice tasks described above. Then, the participants began the transcription tasks and the interface recorded all timestamps for when transcriptions were edited and submitted. After the participants completed all tasks, we collected the timestamp information and computed how long it took to complete each task -- with nine users transcribing nine pages each, we have 81 measurements of transcription speed to be used for quantitatively evaluating the utility of the OCR systems. We also calculated the character error rate (CER) of each transcription with respect to the transcription for the same page by our most experienced participant (a Kwak'wala heritage language learner who is very familiar with the orthography and had transcribed parts of the Hunt/Boas volumes before the user study), and discarded time measurements for transcriptions with CER $\geq 1\%$. Across all 81 transcriptions, only one had an error rate higher than this threshold, and thus, the quantitative analysis below is conducted with 80 time measurements.\footnote{There was no statistical difference between character error rates of from-scratch and corrected transcriptions as well as between participants with and without prior knowledge of Kwak'wala. The participants chosen for the user study had experience in transcription tasks, and all except one transcription were highly accurate (CER $< 1\%$).}

We also obtained qualitative feedback through a short survey that the participants filled out after completing the transcriptions. The survey asked several questions about the experience with the user study, including if the transcribers found specific tasks more difficult than others, whether they preferred typing from scratch or correcting OCR outputs (and which they thought was faster) as well as general feedback on the task and interface. 

\subsection{Quantitative Analysis}
\label{sec:quant}
To quantify the effect of introducing OCR into the transcription process, we analyze the measurements of transcription speed that were collected from the user study tasks. As stated previously, we cannot use the time values directly to make a generalized conclusion because transcription time is not independent of the sources of variability. Instead, we use the statistical technique of Linear Mixed Effects (LME) modeling~\citep{bates2007linear} to describe the relationship between the response variable (the transcription time) and the factors that contribute to variance. The term \ba\ba mixed effects'' refers to a combination of random effects and fixed effects. We have two random effects:
\begin{enumerate}
    \item \textit{transcriber identity}, which can take values from user1 to user9;
    \item \textit{page number}, which can take values from page1 to page9.
\end{enumerate}

\noindent We also have two fixed effects: 
\begin{enumerate}
    \item \textit{transcriber group}, which can either be yes or no indicating prior familiarity with the Kwak’wala language or not;
    \item \textit{task setup}, which can be one of the three setups described above -- baseline, Ocular, or post-correction.
\end{enumerate}

The LME estimation models the transcription time as a function of the above random and fixed effects. Using the estimations, our primary analysis attempts to identify whether the task setup affects transcription time in a statistically significant manner. We additionally look at whether the transcriber group (i.e., whether the participant has prior knowledge of Kwak’wala) plays a role in how fast the user completes tasks.

\input{figures/h1_results}
\paragraph{Does having some form of OCR help reduce transcription time?}
In \autoref{tab:h1results}, we present transcription time estimates from the LME model comparing two settings: (1) the baseline setup which does not use any OCR and the user types the transcription from scratch, and (2) having some form of OCR before the transcription process which the user can correct to produce error-free text (either Ocular or post-correction). As is evident from the results, having some form of OCR greatly improves transcription speed, reducing the time estimate by over 50\% (from 61.65 to 28.21 minutes) and consequently, reducing the manual effort needed to produce an accurate machine-readable version of the documents.

\input{figures/h2_results}
\paragraph{Does post-correction help reduce transcription time beyond using an off-the-shelf OCR tool?}
From the previous results, it is evident that using OCR is beneficial in reducing manual transcription time. We also evaluate whether using the post-correction model is useful or just using an off-the-shelf tool like Ocular is sufficiently useful for transcribers. The LME model estimates for this comparison are in \autoref{tab:h2results}. We see that using post-correction, as proposed in \citet{rijhwani-etal-2021-lexically}, in the transcription pipeline reduces manual correction time by 21\%, indicating its utility to the downstream task of manually correcting the text.

\input{figures/usergroup_results}
\paragraph{Does prior familiarity with Kwak’wala and the Boas script affect transcription time?}
Beyond our primary analysis of the effect of using OCR, we also try to evaluate the extent to which the user’s knowledge of the Kwak’wala language affects the speed of transcription. \autoref{tab:userresults} demonstrates this comparison with results across all three task setups. The estimates show that this factor does play a role with the LME model estimate with a 40\% reduction in transcription time for the group familiar with Kwak’wala. However, the p-value of this estimate is > 0.05, indicating that the result is not statistically significant -- this is likely because only two transcribers in the user study had prior knowledge of the language and more data is needed to draw a statistically significant conclusion.

\subsection{Subjective Feedback}
\label{sec:survey}

After participants completed the transcription tasks, we asked them to fill out a short survey to describe their experience with the task. Note that, to avoid any bias, the participants were not told which OCR setup (Ocular or post-correction) was used for each task. Therefore, the survey focused on understanding whether users observed any differences between typing from scratch or correcting transcriptions, but the questions did not distinguish between the two OCR-based setups. The full list of questions contained in the survey is in \autoref{sec:questions}.

We asked which of the setups led to faster completion of the tasks, and 100\% of the participants perceived that correcting an OCR output was faster than typing the transcription from scratch. Some participants also provided feedback:

\begin{quote}
\textit{\ba\ba Correcting is faster, as there is much less typing involved which requires most of the time'' \\ \hspace*{0pt}\hfill \emph{(user7, from Upwork, not familiar with Kwak'wala)}}
\end{quote}
\vspace{0.05em}
\begin{quote}
\textit{\ba\ba Correcting felt far more efficient!'' \\ \hspace*{0pt}\hfill\emph{(user2, linguistic researcher, familiar with Kwak'wala)}}
\end{quote}
\vspace{0.05em}

However, even though it was slower, two out of the nine participants preferred typing out the text without the aid of an OCR output:

\begin{quote}
\textit{\ba\ba I preferred typing the text from scratch, as searching for any editable text is difficult. You need more effort for editing.'' \\ \hspace*{0pt}\hfill\emph{(user8, from Upwork, not familiar with Kwak'wala)}}
\end{quote}
\vspace{0.05em}

However, the remaining seven transcribers provided strong feedback that correcting OCR outputs was the preferable task setup, for various reasons:

\begin{quote}
\textit{\ba\ba I vastly preferred correcting OCR outputs. It was so much faster, and also required less investment of attention.'' \\ \hspace*{0pt}\hfill\emph{(user2, linguistic researcher, familiar with Kwak'wala)}}
\end{quote}
\vspace{0.05em}
\begin{quote}
\textit{\ba\ba I preferred correcting text - it's much faster. I can spend more mental energy making sure the characters are correct rather than wasting time on transcribing trivially-easy letters.'' \\ \hspace*{0pt}\hfill\emph{(user5, computer science student, not familiar with Kwak'wala)}}
\end{quote}
\vspace{0.05em}
\begin{quote}
\textit{\ba\ba I prefer correcting text because typing from scratch is somehow tricky to follow line by line.'' \\ \hspace*{0pt}\hfill \emph{(user9, from Upwork, not familiar with Kwak'wala)}}
\end{quote}
\vspace{0.05em}

Overall, transcribers participating in the user study identified a reduction in time spent when the OCR outputs were utilized and the majority preferred the task setup not only because of the speed improvement but also because the OCR outputs allowed them to zoom in and fix specific errors rather than spending time on the entire image.

Additionally, we asked participants if any tasks seemed to be easier or more difficult than others. While several described correction as easier than typing from scratch, some transcribers focused on interesting language-specific and document-specific challenges:

\begin{quote}
\textit{\ba\ba A few alphabets were difficult to annotate from the images. For example, it was difficult to differentiate between l and \l{}.'' \\ \hspace*{0pt}\hfill\emph{(user6, from Upwork, not familiar with Kwak'wala)}}
\end{quote}
\vspace{0.05em}
\begin{quote}
\textit{\ba\ba image text was with small fonts.'' \\ \hspace*{0pt}\hfill\emph{(user4, computer science student, not familiar with Kwak'wala)}}
\end{quote}
\vspace{0.05em}
\begin{quote}
\textit{\ba\ba the hardest thing for me was identifying a particular character (\l{}) that is very faint in the original PDF. It is often difficult to tell if a character is \l{} or l. Because I have some knowledge of the language, I relied on that background knowledge at times, but this slowed down the correction process.'' \\ \hspace*{0pt}\hfill (user2, linguistic researcher, familiar with Kwak'wala)}
\end{quote}
\vspace{0.05em}

In giving feedback about the keyboard practice tasks, all participants indicated that the practice task helped them learn the Hunt-Boas orthography and the keystroke mappings. Moreover, 100\% of the participants stated that as they completed more tasks, they became faster at transcription. One participant (user7, from Upwork, not familiar with Kwak'wala) stated \textit{\ba\ba After transcribing a few pages, I became faster at typing with the keyboard and noticing the different accents and letters.''} While the ordering of the tasks was not taken into account in our LME model because of the small amount of data in the current user study, we hope to understand the effect of task order on transcription time in future, larger-scale research.

\section{Conclusion}
In this paper, we evaluate the utility of OCR and post-correction models in a user-centric manner. We conduct a case study on Kwak'wala, an endangered language with a long history of written documentation that is currently not widely accessible to community-based speakers and researchers. With a user study, we highlight the utility of incorporating OCR to make these texts easier to manually transcribe into machine-readable formats. Our statistical analysis shows that OCR can reduce the time taken by a human transcriber in producing an accurate retyping of the texts by over 50\%. While we focus on a single language in this case study, our results demonstrate the immense potential impact that OCR technologies can have on global language documentation and revitalization efforts. Our work, however, is limited in scale and scope -- we do not make statistically significant conclusions on the effect of prior knowledge of the language; whether the order of pages transcribed has an impact on measured time; and the effect of general familiarity with computers and technology. In the future, we hope to conduct a larger-scale evaluation that accounts for these factors; includes transcriptions from a variety of state-of-the-art OCR systems; and expands to more languages, documents, and orthographies.

\section*{Acknowledgements}

This work was supported by the US National Endowment for the Humanities grant PR-276810-21 (\textit{``Unlocking Endangered Language Resources"}) as well as the Government of Canada Social Sciences and Humanities Research Council Insight Development grant GR002807 \textit{\ba\ba \b K'\b an\b k'otła\b x\b ants \b Awi'na\b gwis (Knowing our land)''}. We are very grateful to the transcribers for their participation in the user study and to the reviewers for their helpful feedback.

\bibliography{anthology,custom}
\bibliographystyle{acl_natbib}

\appendix

\section{Appendix}
\subsection{Keyboard for the Boas/Hunt Orthography}
\label{sec:keyboard}
For the user study described in \autoref{sec:userstudy}, we designed a keyboard for the Boas/Hunt orthography to make transcription more efficient.

The keyboard is developed using open-source software Keyman\footnote{\url{https://keyman.com/developer/}} and it maps characters in the Boas orthography to the user's computer keyboard. Keyman also provides an on-screen keyboard to see the mapped layout. We briefly describe the layout and usage of the keyboard below:

\begin{itemize}
    \item Standard English keyboard alphabet and numbers remain in the same position (A-Z, a-z, 0-9) because the Boas orthography uses several Latin script characters.
    \item The special characters, diacritics, and digraphs of the Boas orthography have been assigned to various punctuation keys according to their frequency of use, estimated with a small sample of manually transcribed text (10 pages from~\citet{boas1921}).
    \item All accents are typed after the base character. Examples are shown below:
         \begin{itemize}[label=$\star$]
            \item \"{a} is typed \textit{a then square bracket ]}
            \item k· is typed \textit{k then slash /}
            \item \={o} is typed \textit{o then single quote '}
            \item \^{a} is typed \textit{a then shift + comma ,}
            \item \u{a} is typed \textit{a then shift + period .}
            \item \d{g} is typed \textit{g then shift + square bracket ]}
            \item q´ is typed \textit{q then option (alt key) + 1}
        \end{itemize}
        
    \item Other special characters are:
        \begin{itemize}[label=$\star$]
            \item \textsce{} is assigned to \textit{semicolon ;}
            \item \l{} is assigned to \textit{square bracket [}
            \item \L{} is assigned to \textit{shift + square bracket [}
            \item \textsuperscript{\textepsilon} is assigned to \textit{option (alt key) + e}
            \item \textsuperscript{u} is assigned to \textit{option (alt key) + u}
            \item \textscl{} is assigned to \textit{option (alt key) + l}
        \end{itemize}
    \item All changed punctuation keys can type their original value by holding down the Alt or Option key. For example, to get the original value of the square bracket [, type \textit{Alt + [} (Windows) or \textit{Option + [} (Mac).

\end{itemize}

\subsection{Kwak'wala Transcription: Post-Completion Survey}
\label{sec:questions}

In \autoref{sec:userstudy}, we describe a user study to evaluate the utility of OCR and post-correction models in reducing the time and effort needed for manual transcription. After participants completed transcriptions tasks, we also asked them to fill out a survey to get subjective feedback on their experience with the tasks. Discussion and analysis of the answers from the survey are in \autoref{sec:survey}. We provide a complete list of the questions asked in the survey here:

\begin{enumerate}
    \item Were there specific tasks you found easier or more difficult to annotate?
    \item Did you prefer typing the text from scratch or correcting predictions from a model? Why?
    \item If you are a Kwak'wala language learner, did the annotation help your language learning? How?
    \item Did the practice task help you become familiar with the keyboard?
    \item After annotating a few pages, do you feel like you become faster at annotation?
    \item Which do you feel is faster: typing from scratch or correcting predictions?
    \item Any other feedback or thoughts on the task?
\end{enumerate}

\end{document}

%% file: figures/h1_results.tex
\begin{table}[t]
    \centering
    \begin{tabular}{@{}lcccc@{}}
    \toprule
        Task Setup & Time Est. (\textit{min.}) & $p$-value\\
        \midrule
        Baseline & 61.65 & 3.04e-07 * \\
        With OCR & 28.21 & 4.80e-08 * \\
    \bottomrule
    \end{tabular}
    \caption{Per-page transcription time estimates in minutes from the LME model comparing the baseline, which does not use any OCR, with the task setups that use some form of OCR (either Ocular or post-correction). The time estimate for producing an accurate transcription of a page is reduced by 33.44 minutes when OCR technologies are used beforehand. The $p$-value is $< 0.05$, indicating statistical significance (*).}
    \label{tab:h1results}
\end{table}

%% file: figures/h2_results.tex
\begin{table}[t]
    \centering
    \begin{tabular}{@{}lcccc@{}}
    \toprule
        Task Setup & Time Est. (\textit{min.}) & $p$-value\\
        \midrule
        Ocular & 31.67 & 2.55e-05 * \\
        Post-correction & 24.98 & 0.0121 * \\
    \bottomrule
    \end{tabular}
    \caption{Per-page transcription time estimates in minutes from the LME model comparing task setups using an off-the-shelf OCR system (Ocular) with an OCR post-correction method. The time estimate is reduced by 6.69 minutes for a page, indicating the utility of post-correction to downstream users over using Ocular. The $p$-value is $< 0.05$, indicating statistical significance (*).}
    \label{tab:h2results}
\end{table}

%% file: figures/usergroup_results.tex
\begin{table}[t]
    \centering
    \begin{tabular}{@{}lcccc@{}}
    \toprule
        Group & Time Est. (\textit{min.}) & $p$-value\\
        \midrule
        Not familiar & 43.60 & 8.12e-05 * \\
        Familiar & 25.74 & 0.228 \\
    \bottomrule
    \end{tabular}
    \caption{Per-page transcription time estimates in minutes from the LME model comparing transcribers that had prior familiarity with Kwak'wala with those that did not. The time estimate is reduced by 17.86 minutes for a page when the user is familiar with Kwak'wala, indicating that target knowledge language might be useful to have in image transcription tasks. The $p$-value is $> 0.05$ for the estimate, which indicates that it is not statistically significant, likely because we only had two users that were familiar with the language.}
    \label{tab:userresults}
\end{table}

%% file: acl_latex.bbl
\begin{thebibliography}{25}
\expandafter\ifx\csname natexlab\endcsname\relax\def\natexlab#1{#1}\fi

\bibitem[{Bahdanau et~al.(2015)Bahdanau, Cho, and Bengio}]{bahdanau2015neural}
Dzmitry Bahdanau, Kyunghyun Cho, and Yoshua Bengio. 2015.
\newblock Neural machine translation by jointly learning to align and
  translate.
\newblock In \emph{3rd International Conference on Learning Representations,
  ICLR 2015}.

\bibitem[{Bates(2007)}]{bates2007linear}
Douglas Bates. 2007.
\newblock Linear mixed model implementation in lme4.
\newblock \emph{Manuscript, University of Wisconsin}, 15.

\bibitem[{Berg-Kirkpatrick et~al.(2013)Berg-Kirkpatrick, Durrett, and
  Klein}]{berg-kirkpatrick-etal-2013-unsupervised}
Taylor Berg-Kirkpatrick, Greg Durrett, and Dan Klein. 2013.
\newblock \href {https://aclanthology.org/P13-1021} {Unsupervised transcription
  of historical documents}.
\newblock In \emph{Proceedings of the 51st Annual Meeting of the Association
  for Computational Linguistics (Volume 1: Long Papers)}, pages 207--217,
  Sofia, Bulgaria. Association for Computational Linguistics.

\bibitem[{Boas(1897)}]{boas1897social}
Franz Boas. 1897.
\newblock \emph{The Social Organization and the Secret Societies of the
  Kwakiutl Indians: Smithsonian Institution. United States National Museum. By
  Franz Boas. With 51 Plates}.
\newblock Washington: G.P.O.

\bibitem[{Boas(1911)}]{boas1911}
Franz Boas. 1911.
\newblock \emph{“Kwakiutl.” Pp. 423–557 in Handbook of American Indian
  Languages, vol. 40.1, Bureau of American Ethnology Bulletin, edited by Franz
  Boas.}
\newblock Washington: G.P.O.

\bibitem[{Boas(1934)}]{boas1934}
Franz Boas. 1934.
\newblock \emph{Geographical Names of the Kwakiutl Indians.}
\newblock New York: Columbia University Press.

\bibitem[{Boas and Hunt(1902)}]{boas1902}
Franz Boas and George Hunt. 1902.
\newblock \emph{Kwakiutl Texts.}
\newblock Leiden, New York: E.J. Brill; G.E. Stechert \& Co.

\bibitem[{Boas and Hunt(1921)}]{boas1921}
Franz Boas and George Hunt. 1921.
\newblock \emph{Ethnology of the Kwakiutl: Based on Data Collected by George
  Hunt}.
\newblock Washington: G.P.O.

\bibitem[{Dean and Voss(1999)}]{dean1999design}
Angela Dean and Daniel Voss. 1999.
\newblock \emph{Design and analysis of experiments}.
\newblock Springer.

\bibitem[{Disbray et~al.(2022)Disbray, Foley, Rijhwani, and Mistica}]{pintupi}
Samantha Disbray, Ben Foley, Shruti Rijhwani, and Meladel Mistica. 2022.
\newblock Reading it right: A case study in pintupi-luritja.
\newblock In \emph{Digital Approaches to Multilingual Text Analysis}.

\bibitem[{Dong and Smith(2018)}]{dong-smith-2018-multi}
Rui Dong and David Smith. 2018.
\newblock \href {https://doi.org/10.18653/v1/P18-1220} {Multi-input attention
  for unsupervised {OCR} correction}.
\newblock In \emph{Proceedings of the 56th Annual Meeting of the Association
  for Computational Linguistics (Volume 1: Long Papers)}, pages 2363--2372,
  Melbourne, Australia. Association for Computational Linguistics.

\bibitem[{Garrette et~al.(2015)Garrette, Alpert-Abrams, Berg-Kirkpatrick, and
  Klein}]{garrette-etal-2015-unsupervised}
Dan Garrette, Hannah Alpert-Abrams, Taylor Berg-Kirkpatrick, and Dan Klein.
  2015.
\newblock \href {https://doi.org/10.3115/v1/N15-1109} {Unsupervised
  code-switching for multilingual historical document transcription}.
\newblock In \emph{Proceedings of the 2015 Conference of the North {A}merican
  Chapter of the Association for Computational Linguistics: Human Language
  Technologies}, pages 1036--1041, Denver, Colorado. Association for
  Computational Linguistics.

\bibitem[{Gaspari et~al.(2014)Gaspari, Toral, Naskar, Groves, and
  Way}]{gaspari-etal-2014-perception}
Federico Gaspari, Antonio Toral, Sudip~Kumar Naskar, Declan Groves, and Andy
  Way. 2014.
\newblock \href {https://aclanthology.org/2014.amta-wptp.5} {Perception vs.
  reality: measuring machine translation post-editing productivity}.
\newblock In \emph{Proceedings of the 11th Conference of the Association for
  Machine Translation in the Americas}, pages 60--72, Vancouver, Canada.
  Association for Machine Translation in the Americas.

\bibitem[{Grenoble and Whaley(2005)}]{grenoble2005saving}
Lenore~A Grenoble and Lindsay~J Whaley. 2005.
\newblock \emph{Saving languages: An introduction to language revitalization}.
\newblock Cambridge University Press.

\bibitem[{Himmelmann(1998)}]{himmelmann1998documentary}
Nikolaus~P Himmelmann. 1998.
\newblock Documentary and descriptive linguistics.

\bibitem[{Kettunen et~al.(2022)Kettunen, Keskustalo, Kumpulainen,
  P{\"a}{\"a}kk{\"o}nen, and Rautiainen}]{kettunen2022ocr}
Kimmo Kettunen, Heikki Keskustalo, Sanna Kumpulainen, Tuula
  P{\"a}{\"a}kk{\"o}nen, and Juha Rautiainen. 2022.
\newblock Ocr quality affects perceived usefulness of historical newspaper
  clippings--a user study.
\newblock \emph{arXiv preprint arXiv:2203.03557}.

\bibitem[{Kolak and Resnik(2005)}]{kolak-resnik-2005-ocr}
Okan Kolak and Philip Resnik. 2005.
\newblock \href {https://aclanthology.org/H05-1109} {{OCR} post-processing for
  low density languages}.
\newblock In \emph{Proceedings of Human Language Technology Conference and
  Conference on Empirical Methods in Natural Language Processing}, pages
  867--874, Vancouver, British Columbia, Canada. Association for Computational
  Linguistics.

\bibitem[{Koponen(2016)}]{koponen2016machine}
Maarit Koponen. 2016.
\newblock Is machine translation post-editing worth the effort? a survey of
  research into post-editing and effort.
\newblock \emph{The Journal of Specialised Translation}, 25:131--148.

\bibitem[{Krishna et~al.(2018)Krishna, Majumder, Bhat, and
  Goyal}]{krishna-etal-2018-upcycle}
Amrith Krishna, Bodhisattwa~P. Majumder, Rajesh Bhat, and Pawan Goyal. 2018.
\newblock \href {https://doi.org/10.18653/v1/K18-1034} {Upcycle your {OCR}:
  Reusing {OCR}s for post-{OCR} text correction in {R}omanised {S}anskrit}.
\newblock In \emph{Proceedings of the 22nd Conference on Computational Natural
  Language Learning}, pages 345--355, Brussels, Belgium. Association for
  Computational Linguistics.

\bibitem[{Lawson(2004)}]{Lawson_2004}
Kimberley~L. Lawson. 2004.
\newblock \href {https://doi.org/http://dx.doi.org/10.14288/1.0091657}
  {\emph{Precious fragments: First Nations materials in archives, libraries and
  museums}}.
\newblock Ph.D. thesis, University of British Columbia.

\bibitem[{Rijhwani et~al.(2020)Rijhwani, Anastasopoulos, and
  Neubig}]{rijhwani-etal-2020-ocr}
Shruti Rijhwani, Antonios Anastasopoulos, and Graham Neubig. 2020.
\newblock \href {https://doi.org/10.18653/v1/2020.emnlp-main.478} {{OCR} {P}ost
  {C}orrection for {E}ndangered {L}anguage {T}exts}.
\newblock In \emph{Proceedings of the 2020 Conference on Empirical Methods in
  Natural Language Processing (EMNLP)}, pages 5931--5942, Online. Association
  for Computational Linguistics.

\bibitem[{Rijhwani et~al.(2021)Rijhwani, Rosenblum, Anastasopoulos, and
  Neubig}]{rijhwani-etal-2021-lexically}
Shruti Rijhwani, Daisy Rosenblum, Antonios Anastasopoulos, and Graham Neubig.
  2021.
\newblock \href {https://doi.org/10.1162/tacl_a_00427} {Lexically aware
  semi-supervised learning for {OCR} post-correction}.
\newblock \emph{Transactions of the Association for Computational Linguistics},
  9:1285--1302.

\bibitem[{Schulz and Kuhn(2017)}]{schulz-kuhn-2017-multi}
Sarah Schulz and Jonas Kuhn. 2017.
\newblock \href {https://doi.org/10.18653/v1/D17-1288} {Multi-modular
  domain-tailored {OCR} post-correction}.
\newblock In \emph{Proceedings of the 2017 Conference on Empirical Methods in
  Natural Language Processing}, pages 2716--2726, Copenhagen, Denmark.
  Association for Computational Linguistics.

\bibitem[{Specia and Farzindar(2010)}]{specia2010estimating}
Lucia Specia and Atefeh Farzindar. 2010.
\newblock Estimating machine translation post-editing effort with hter.
\newblock In \emph{Proceedings of the Second Joint EM+/CNGL Workshop: Bringing
  MT to the User: Research on Integrating MT in the Translation Industry},
  pages 33--43.

\bibitem[{Tjuatja et~al.(2021)Tjuatja, Rijhwani, and
  Neubig}]{tjuatjaexplorations}
Lindia Tjuatja, Shruti Rijhwani, and Graham Neubig. 2021.
\newblock Explorations in transfer learning for ocr post-correction.
\newblock In \emph{Fifth Widening Natural Language Processing Workshop
  (WiNLP)}.

\end{thebibliography}
